\documentclass[letterpaper]{article}
\usepackage{aaai}
\usepackage{times}
\usepackage{helvet}
\usepackage{courier}
\usepackage{xcolor}
\usepackage{amsmath,amssymb,amsfonts}
\usepackage[ruled, lined, linesnumbered]{algorithm2e}
\usepackage{tikz}
\usetikzlibrary{er,positioning}
\usepackage{graphicx}
\usepackage{textcomp}
\usepackage{xcolor}
\usepackage{numprint}
\npthousandsep{,}
\usepackage{multirow}
\usepackage{hyperref}
\usepackage{bbm}
\usepackage[inline, shortlabels]{enumitem}
\usepackage{todonotes}

\begin{document}
%
\title{Designing Emergency Response Pipelines : Lessons and Challenges}
\author{Ayan Mukhopadhyay\\
{Stanford University}\\
ayanmukh@stanford.edu
\And
Geoffrey Pettet\\
{Vanderbilt University}\\
geoffrey.a.pettet@vanderbilt.edu
\And
Mykel Kochenderfer\\
{Stanford University}\\
mykel@stanford.edu
\And
Abhishek Dubey\\
{Vanderbilt University}\\
abhishek.dubey@vanderbilt.edu}
\maketitle
\begin{abstract}
\begin{quote}
Emergency response to incidents such as accidents, crimes, and fires is a major problem faced by communities. Emergency response management comprises of several stages and sub-problems like forecasting, resource allocation, and dispatch. The design of principled approaches to tackle each problem is necessary to create efficient emergency response management (ERM) pipelines. Over the last six years, we have worked with several first responder organizations to design ERM pipelines. In this paper, we highlight some of the challenges that we have identified and lessons that we have learned through our experience in this domain. Such challenges are particularly relevant for practitioners and researchers, and are important considerations even in the design of response strategies to mitigate disasters like floods and earthquakes. 
\end{quote}
\end{abstract}

\section{Introduction}
Emergency response to incidents such as accidents, medical calls, urban crimes, poaching, and fires is one of the most pressing problems faced by communities across the globe. Emergency response must be fast and efficient in order to minimize loss of lives~\cite{jaldell2017important,jaldell2014time}. Significant attention in the last several decades has been devoted to studying emergency incidents and response. The broader goal of designing ERM systems is to enable communities to deal with emergency response in a manner that is principled, proactive, and efficient. ERM systems based on data-driven models can help reduce both human and financial losses. Insights from principled approaches can also be used to improve policies and safety measures. Although such models are increasingly being adopted by government agencies, emergency incidents still cause thousands of deaths and injuries and also result in losses worth more than billions of dollars directly or indirectly each year~\cite{crimeUS}.

ERM can be divided into five major components: 1) mitigation, 2) preparedness, 3) detection, 4) response, and 5) recovery. The stages of ERM systems are heavily interlinked~\cite{mukhopadhyay2020review}. Mitigation involves sustained efforts to reduce long-term risks to people and property. It also involves creating forecasting models to understand the spatial and temporal characteristics of incidents. Preparedness involves creating policy and allocating resources that enable emergency response management. The third phase seeks to use automated techniques to detect incidents as they happen in order to expedite response. The dispatch phase, the most critical phase in the field, involves responding to incidents when they occur. Finally, the recovery phase seeks to provide support and sustenance to communities and individuals affected by the emergency. While most of the prior work in ERM has studied these problems independently, these stages are inter-linked. Frequently, the output of one stage serves as the input for another. For example, predictive models learned in the \textit{preparedness} stage are used in planning \textit{response} strategies. Therefore, it is crucial that ERM pipelines are designed such that intricate interdependencies are considered. In this paper, we highlight some of the challenges that we have faced while designing ERM pipelines and lessons that we have learned through the process. A large portion of the insights have come from our colleagues at the Nashville Fire Department and the Tennessee Department of Transportation, who have provided invaluable domain expertise to us. 

\section{Challenges and Lessons}

While ERM forms a crucial component of the cities and governments, designing and deploying principled approaches to ERM is challenging. We consider the following to be the main challenges in the design and deployment of ERM pipelines.

\begin{enumerate}
    \item \textbf{How to forecast incident occurrence?} It has been noted that emergency incidents are generally difficult to predict due to the inherent random nature of such incidents and spatially varying factors~\cite{shankar1995effect,mukhopadhyayAAMAS17}. Incidents are highly sporadic, making it particularly difficult to learn models of incident occurrence. For example, well-known regression models such as Poisson regression and negative binomial regression have been shown to perform poorly on accident data due to the prevalence of zero counts~\cite{mukhopadhyay2020review}. Incident data has also been shown to be particularly sensitive to the scale of spatial and temporal resolutions, thereby making it difficult to perform meaningful inference. There are several approaches that have shown to alleviate these concerns. First, identifying clusters of incidents (both spatial and non-spatial) has shown to balance model variance and spatial heterogeneity particularly well~\cite{sasidharan2015exploring,mukhopadhyayAAMAS17}. Second, dual state models like zero-inflated Poisson models can be used to address the issue of a high number of zero counts in data~\cite{qin2004selecting}.
    
    \item \textbf{When to optimize?} Arguably, the most important component of ERM pipelines is to dispatch responders when incidents occur. While resource allocation and dispatch to emergency incidents evolve in highly uncertain and dynamic environments, the expectation is that response is very timely~\cite{felder2002spatial,mukhopadhyayAAMAS18}. Approaches to optimize dispatch typically focus on decision-making \textit{after} an incident occurs~\cite{toro2013joint,mukhopadhyayAAMAS18,keneally2016markov}. However, our conversations and collaborations with first responders revealed that there is limited applicability of such approaches in practice due to two important reasons. First, response to incidents occurs almost instantaneously after a report is received. Although optimizing dispatch can minimize response times in the long run, time spent to optimize dispatch \textit{after} incidents occur is perceived as costly in the field. Second, it is almost impossible to judge the severity of an incident from a call for service. Consequently, it is imperative for first responders to follow a \textit{greedy} strategy and dispatch the closest available responder to incidents. An alternative approach is to periodically optimize the spatial distribution of responders \textit{between} incidents. ~\citeauthor{pettet2020algorithmic}~\cite{pettet2020algorithmic} introduced this idea recently for emergency response. While there are challenges with respect to scalability of such an approach, planning between incidents is much more applicable in the field, as it does not violate constraints under which first responder operate. 

    \item \textbf{How to model communication?} Approaches to tackle emergency response typically assume that the agents can observe the world completely and communicate with centralized servers and each other. This assumption is usually satisfied in practice. However, in scenarios that involve natural disasters (like floods, earthquakes, wildfires, etc.) communication mechanisms can break down and power failures are common. In such scenarios, it is important that agents can optimize response based on information gathered locally. One way to approach such a problem is to design distributed approaches for ERM, in which agents can optimize their own decisions~\cite{pettet2020algorithmic}. This is feasible since modern agents (ambulances, for example) are equipped with laptops. While distributed approaches perform worse than their centralized counterparts, they provide the benefit of performance in scenarios where communication is challenging. 
    
    \item \textbf{How to model the environment?} A problem in designing approaches to ERM is that environmental conditions under which response takes place is dynamic. Consider a decision-theoretic model for dispatching responders. For example, see the semi-Markov decision process formulation to minimize expected response times to accidents ~\cite{mukhopadhyayAAMAS18}. An approach to solve such large-scale decision problems is to use a simulator to find a policy that picks the optimal action given the state of the world. However, in urban areas, events like road closures, constructions, or increased traffic due to a public gathering can drastically alter the distribution of incidents. Further, ambulances can be unavailable due to breakdowns or maintenance. In such cases, it is crucial that the actual state of the environment is taken into account while creating allocation and dispatch decisions. One approach is to create high-fidelity models of covariates such as traffic and weather. Such models can then be used in online forecasting models, that can accommodate incoming streams of updated information~\cite{MukhopadhyayICCPS}. Similarly, decision-theoretic approaches that can quickly compute promising actions for the current state of the world can be more valuable in emergency response than approaches that find policies for the entire state space of the problem~\cite{pettet2020algorithmic}. 
\end{enumerate}

The principled design of ERM systems is an important problem faced by communities. As smart and connected communities evolve, they present both opportunities and challenges to manage ERM systems. In this short paper, we highlight common problems and lessons learned through our experience in designing ERM systems over the last few years. 

\section{Acknowledgement}

We would like to acknowledge the National Science Foundation and the Center of Automotive Research at Stanford for funding this research. We would also like to thank the Nashville Fire Department (NFD) and the Tennessee Department of Transportation (TDOT) for collaborating with us and providing invaluable knowledge about the intricacies of emergency response.

{\small
\bibliographystyle{aaai}
\bibliography{references}
}

\end{document}